\documentclass[11pt]{article}

\usepackage{acl}
\usepackage{fancyhdr}
\usepackage{graphicx}
\usepackage{xifthen}

\newboolean{showlogo}
\setboolean{showlogo}{true}   

\newcommand{\logopath}{./figures/qwen-BU.png}

\fancypagestyle{firststyle}{%
  \fancyhf{}%
  \fancyhead[L]{%
    \ifthenelse{\boolean{showlogo}}{%
      \includegraphics[width=115pt]{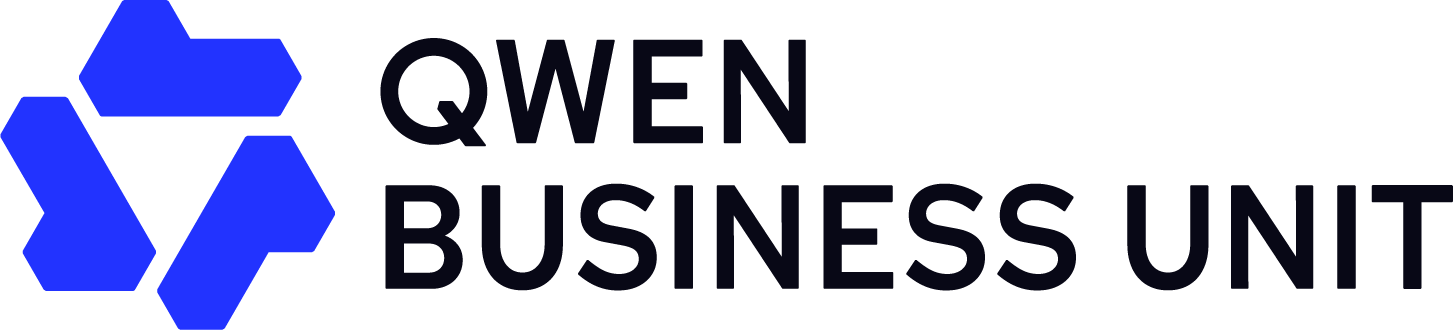}%
    }{}%
  }%
}

\usepackage{times}
\usepackage{latexsym}
\usepackage[T1]{fontenc}
\usepackage[utf8]{inputenc}
\usepackage{microtype}

\usepackage{inconsolata}
\usepackage{graphicx}
\usepackage{booktabs}
\usepackage{multirow}
\usepackage{amsmath}
\usepackage{amssymb}
\usepackage{tcolorbox}
\tcbuselibrary{listings, breakable}

\usepackage{listings}

\definecolor{codegreen}{rgb}{0,0.6,0}
\definecolor{codegray}{rgb}{0.5,0.5,0.5}
\definecolor{codepurple}{rgb}{0.58,0,0.82}
\lstdefinestyle{mystyle}{
    commentstyle=\color{codegreen},
    keywordstyle=\color{magenta},
    numberstyle=\tiny\color{codegray},
    stringstyle=\color{codepurple},
    basicstyle=\ttfamily\footnotesize,
    breakatwhitespace=false,         
    breaklines=true,                 
    captionpos=b,                    
    keepspaces=true,                 
    numbers=left,                    
    numbersep=5pt,                  
    showspaces=false,                
    showstringspaces=false,
    showtabs=false,                  
    tabsize=2
}
\title{\textbf{BoRP}: Bootstrapped Regression Probing for \\ Scalable and Human-Aligned LLM Evaluation}

\author{
  \textbf{Peng Sun}\thanks{Corresponding author.},
  \textbf{Xiangyu Zhang},
  \textbf{Duan Wu} \\[0.3em]
  \textbf{Lu Tan},
  \textbf{Jian Lin},
  \textbf{He Yang},
  \textbf{Qi Qian},
  \textbf{Yikai Wang}\\[0.5em]
  Qwen Business Unit of Alibaba \\
  \texttt{sp80287@alibaba-inc.com}
}

\begin{document}
\maketitle


\begin{abstract}
Session-level satisfaction scoring is critical for the iterative development of open-ended conversational AI, yet existing options are unsatisfactory: explicit feedback is sparse, implicit heuristics are ambiguous, and generative LLM-as-a-Judge approaches suffer from central-tendency bias, output-protocol sensitivity, and prohibitive decoding cost at full traffic.
We introduce \textbf{BoRP} (Bootstrapped Regression Probing), which takes a different route: \emph{rather than asking an LLM to write out a judgment, BoRP reads the judgment directly out of the base model's hidden states}. By extracting continuous 1--5 scores from the latent space via supervised regression, BoRP eliminates the need for token decoding. Furthermore, it is supported by a cold-start bootstrapping pipeline that distills an evaluation rubric from unlabeled traffic without massive manual curation.
This ``read, don't write'' paradigm fundamentally reshapes the evaluation trade-off space, yielding state-of-the-art human alignment while slashing inference costs to a fraction of generative baselines. Furthermore, its unique backbone evolvability allows seamless model upgrades with minimal retraining, finally making full-traffic satisfaction monitoring and highly sensitive A/B testing a practical reality.
\end{abstract}


\begin{figure*}[t]
    \centering
    \includegraphics[width=\textwidth]{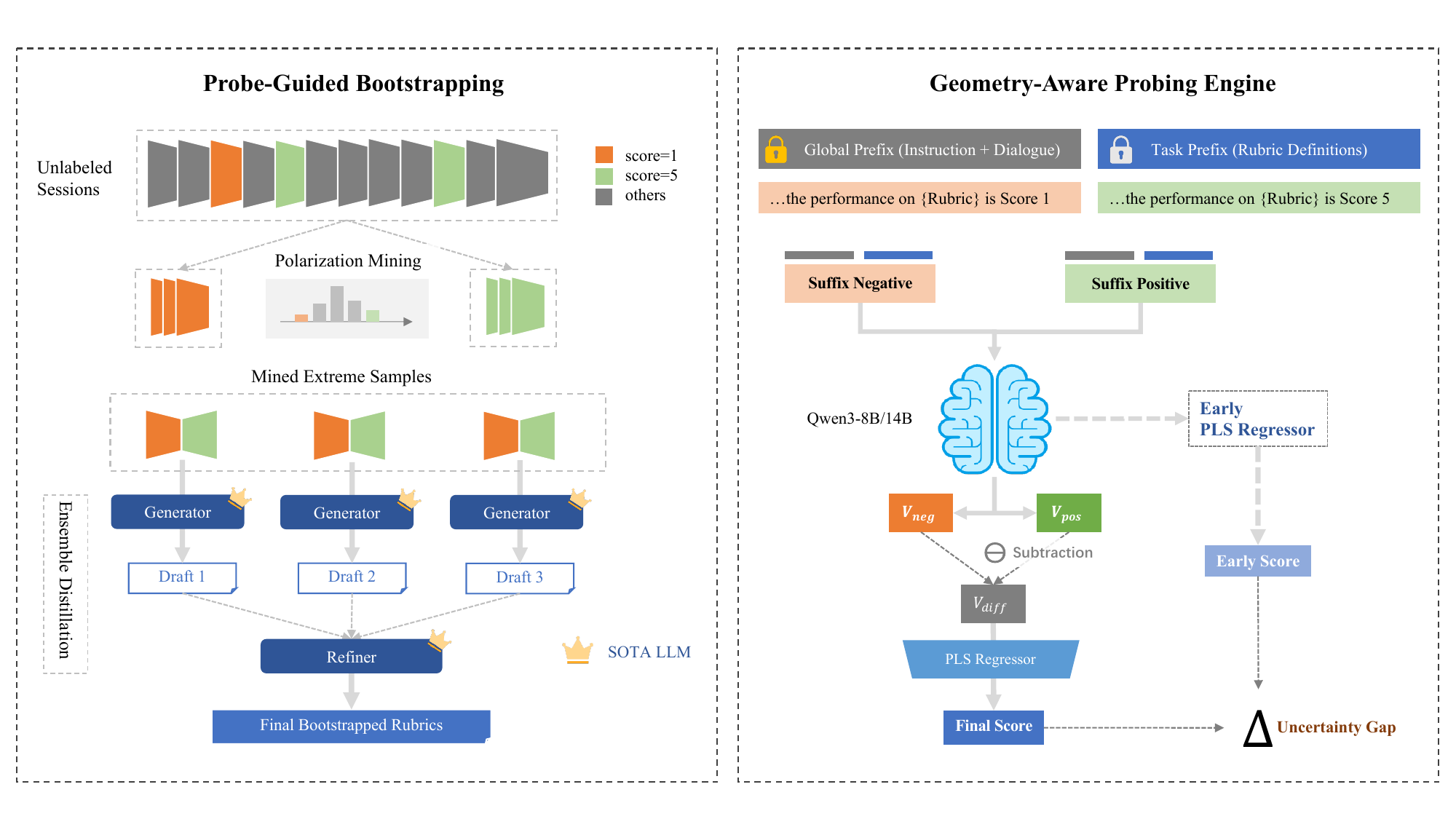} 
    \caption{\textbf{The BoRP Framework.} \textbf{(Left) Probe-Guided Bootstrapping:} Utilizing the Polarization Index (PI) to mine extreme samples from unlabeled traffic for cold-start rubric synthesis. \textbf{(Right) Probing \& Inference Engine:} Extracting contrastive hidden states via suffix-only prompting (optimizing KV-cache reuse). A lightweight PLS head maps representations to continuous scores, while layer-wise discrepancies quantify uncertainty for tiered evaluation.}
    \label{fig:teaser}
\end{figure*}

\section{Introduction}

LLM-powered conversational agents have shifted to open-ended interactions, complicating the quantification of user satisfaction. Heuristic metrics (Session Duration, Turn Counts) are ambiguous \citep{deriu2021survey}---prolonged engagement can reflect immersion or confusion---while posterior signals such as retention lag too far behind to drive rapid iteration. This \emph{Metric Gap} calls for a framework that captures \emph{immediate}, session-level satisfaction at full traffic to power sensitive A/B testing \citep{kohavi2020trustworthy}.

Existing approaches, however, exhibit distinct structural limitations. Manual evaluation cannot reach the sample sizes A/B testing demands. Generative LLM-as-a-Judge incurs an ``alignment tax'' (central-tendency bias); specialized fine-tuned judges \citep{kim2024prometheus, zhu2025judgelm, chen2025judgelrm} demand $>$100K curated samples per backbone and remain prompt-protocol sensitive \citep{wang2025improving}. Conventional probing \citep{burns2022discovering}, though efficient, settles for coarse classification. Across all three, raising fidelity raises cost.

To resolve this, we propose \textbf{BoRP}. Our framework shifts the paradigm: \emph{rather than articulating a textual judgment, we extract the satisfaction signal directly from the LLM's internal hidden states}. We isolate this direction via contrastive prompts, map it to a continuous 1--5 score using a lightweight PLS regression head, and eliminate the cold-start burden via a polarization-driven bootstrapping pipeline that distills rubrics from unlabeled traffic. This ``read, don't write'' paradigm yields three coupled advantages:
(i)~Stronger human alignment---achieving a Krippendorff's alpha (K-$\alpha$) of 0.80 vs.\ 0.73 for a leading general-purpose LLM, and 0.69 vs.\ 0.49 when compared to a specialized generative judge fine-tuned on the same backbone, with no alignment tax, verbosity/position bias, or output-protocol sensitivity;
(ii)~$>$30$\times$ lower inference cost---one A100 sustains $\sim$1.4M sessions/day at $\sim$\$4 per 100K sessions via KV-cache reuse and suffix-only probing;
and (iii)~backbone evolvability---swapping the base model retrains only the lightweight head on the same 700 expert annotations, with no $>$100K re-curation or re-SFT.

\section{Related Work}
\label{sec:related_work}

\paragraph{LLM-as-a-Judge \& Specialized Models}

In LLM evaluation, particularly for open-ended multi-turn dialogue, the field has shifted from n-gram metrics to prompting GPT-4 \citep{zheng2024judging}, and more recently to fine-tuning specialized judges---most notably the Prometheus line \citep{kim2024prometheus, kim2024prometheus2, pombal2025mprometheus}, JudgeLM \citep{zhu2025judgelm}, and JudgeLRM \citep{chen2025judgelrm}---which train on $>$100K curated samples to produce rubric-grounded verbal feedback.

While effective, these generative judges share three structural limitations: (i) the $>$100K supervision requirement is prohibitive when ground truth must come from domain experts; (ii) being generative, they suffer from position bias \citep{wang2023large}, verbosity bias \citep{saito2023verbosity}, and calibration failures \citep{lee2025reporting, wang2025improving}; and (iii) they cannot track the foundation-model capability frontier without re-synthesizing data and redoing full SFT per new backbone. BoRP takes the opposite route: discriminative regression on hidden states requires only $\sim$700 expert-labeled samples, avoids generation-side biases, and adapts to a new backbone by retraining only the lightweight regression head.

\paragraph{Probing \& Rubric Generation}

Linear probing has a long tradition in interpretability \citep{alain2016understanding} but has stayed predominantly binary---truthfulness detection \citep{burns2022discovering}, or, closest to our setting, classifying probes for pairwise preference extraction \citep{maiya2025improving}, which establishes that hidden-state readout can outperform generative judges on relative preference. Recent auto-rubric methods \citep{liu2026openrubrics, xie2025autorubric} rely on prompting. BoRP extends this line in two directions: from binary/pairwise to fine-grained \emph{absolute regression} via PLS, and from given-rubric to a polarization-driven bootstrapping pipeline that mines rubric-defining extreme cases from unlabeled traffic.
\section{Methodology}
\label{sec:method}

We instantiate BoRP as a fully discriminative pipeline: each session's hidden state is projected through a lightweight regression head onto a calibrated 1--5 score, with no token-level decoding required at inference.

Realizing this idea requires resolving three coupled challenges: no off-the-shelf rubric exists for a new business scenario, extreme cases are sparse and entangled with topic noise in the latent space, and full-traffic scoring imposes hard throughput budgets. As illustrated in Figure~\ref{fig:teaser}, we address them through three components: Probe-Guided Bootstrapping (Sec.~\ref{sec:bootstrapping}) for cold-start rubric synthesis, the Geometry-Aware Probing Engine (Sec.~\ref{sec:regression}) for sample-efficient regression on hidden states, and the High-Throughput Inference Engine (Sec.~\ref{sec:system}) for full-traffic deployment. This hidden-state-centric design also confers two structural advantages: \emph{backbone evolvability}---adapting to a new base model only requires retraining the lightweight regression head on the same samples; and \emph{protocol-failure avoidance}---producing no verbal feedback sidesteps the output-protocol sensitivity of generative judges (Appendix~\ref{app:gen_baselines}).

Throughout, the supervisory roles are strictly separated: the teacher LLM is invoked once to distill a rubric from polarization-mined extreme cases (Sec.~\ref{sec:bootstrapping}); all $\sim$700 session-level 1--5 labels are then produced by human experts against this rubric (Sec.~\ref{sec:regression}). The teacher is never used at training or inference time.

\begin{figure*}[t!]
    \centering
    \begin{minipage}{0.32\textwidth}
        \centering
        \includegraphics[width=\textwidth]{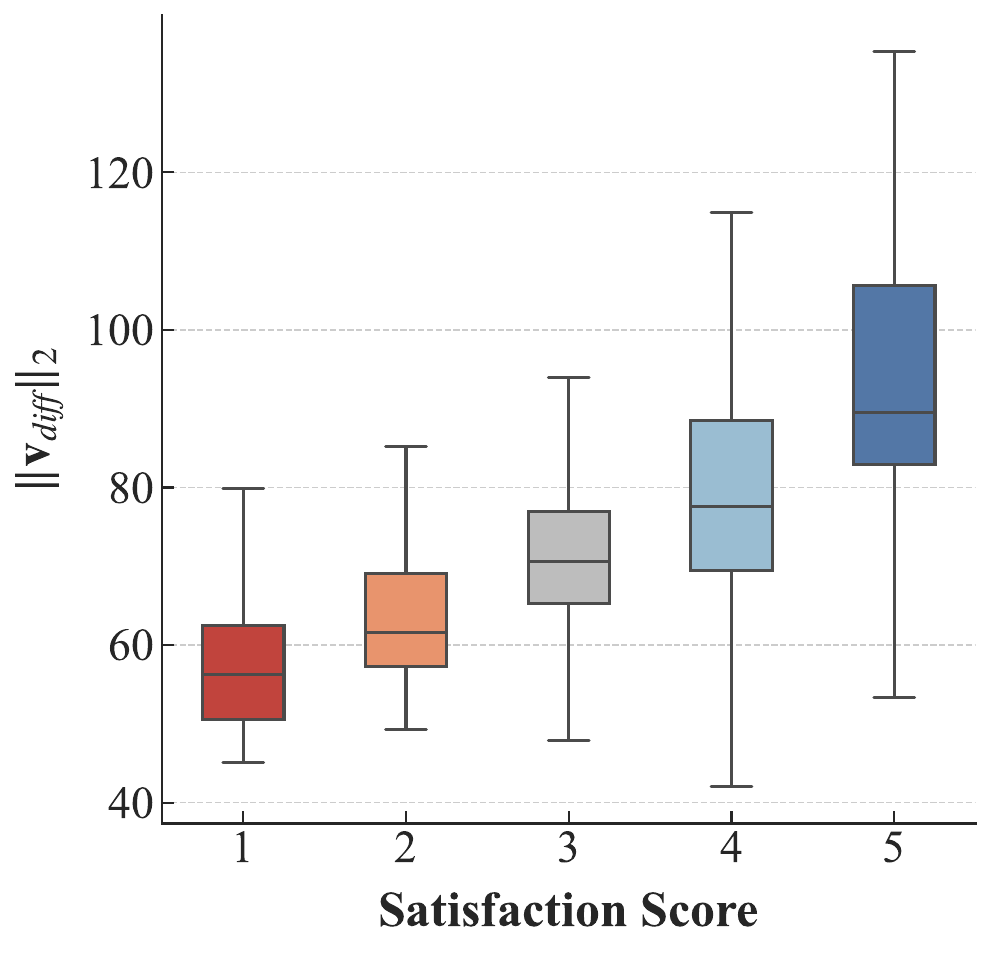}
        \caption*{(a) Norm vs. Score}
    \end{minipage}
    \hfill
    \begin{minipage}{0.32\textwidth}
        \centering
        \includegraphics[width=\textwidth]{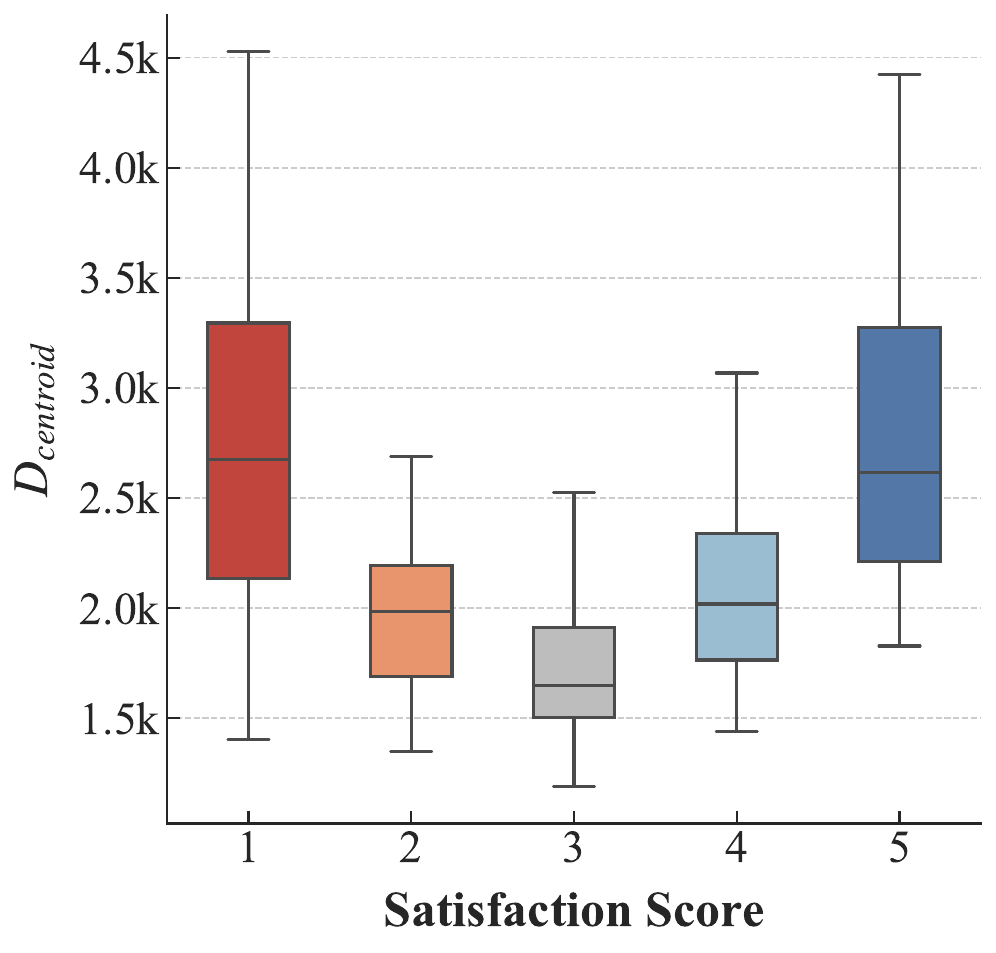}
        \caption*{(b) Distance vs. Score}
    \end{minipage}
    \hfill
    \begin{minipage}{0.32\textwidth}
        \centering
        \includegraphics[width=\textwidth]{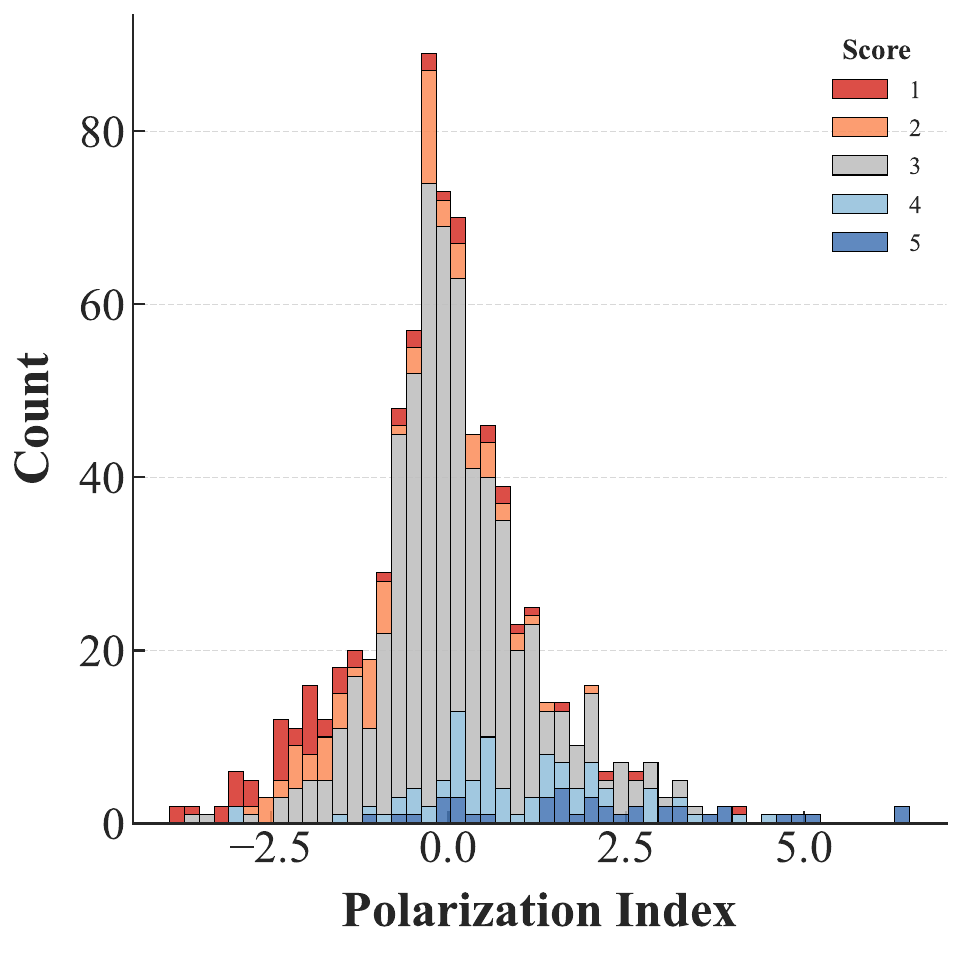}
        \caption*{(c) PI Distribution}
    \end{minipage}
    
    \caption{\textbf{Geometric Analysis.} (a) Vector norm correlates with intensity. (b) Distance to centroid indicates outlierness. (c) PI effectively polarizes extremes to the tails.}
    \label{fig:geometry}
\end{figure*}

\subsection{Preliminaries: Contrastive Representation}
\label{sec:preliminaries}

Following \citet{burns2022discovering}, we extract representations from the model's internal states. For a dialogue context $H$ and query $U$, we construct a positive prompt $P_{\text{pos}}$ and a negative prompt $P_{\text{neg}}$. Let $h_L^{(T)}(x) \in \mathbb{R}^d$ denote the extracted hidden state of the final token for input $x$. We compute the difference vector:
\begin{equation}
\label{eq:v_diff}
\mathbf{v}_{\text{diff}} = h_L^{(T)}(P_{\text{pos}}) - h_L^{(T)}(P_{\text{neg}})
\end{equation}
This difference vector $\mathbf{v}_{\text{diff}}$ isolates the satisfaction signal: because both prompts append to the shared context $(H, U)$, the representational contribution of topic and syntax is common-mode and cancels out in the subtraction, leaving only the directional shift induced by the polar prompts.

\subsection{Probe-Guided Criteria Bootstrapping}
\label{sec:bootstrapping}

A discriminative 1--5 rubric is the precondition for everything that follows---yet writing one by hand for each new business scenario is prohibitively expensive, and the polar 1- and 5-point cases that anchor it are ``needles in a haystack'' inside average-skewed industrial traffic. We therefore invert the usual order (Figure~\ref{fig:teaser} Left): rather than writing a rubric and then searching for examples, we first mine the examples geometrically from unlabeled traffic, then have a teacher LLM induce the rubric from them.

\paragraph{Blind Probing}
Lacking specific rubrics, we extract a coarse $\mathbf{v}_{\text{diff}}$ via Eq.~\eqref{eq:v_diff} using generic semantic prompts (see Figure~\ref{fig:prompt_blind_probe} in Appendix~\ref{app:prompts}), capturing the model's innate intuition about quality.

\paragraph{Polarization Mining}
Effective mining requires locating extreme cases in the latent space. Exploratory analysis (Figure~\ref{fig:geometry}a-b) reveals two geometric signatures of $\mathbf{v}_{\text{diff}}$ for extreme scores: \textbf{Intensity} (the Euclidean norm $\|\mathbf{v}_{\text{diff}}\|_2$ correlates positively with satisfaction) and \textbf{Outlierness} (the Manhattan distance to the centroid $\mathbf{c}$ follows a U-shaped distribution). Fusing these, we formulate the \textbf{Polarization Index (PI)}:

\begin{equation}
\label{eq:pi}
\mathrm{PI} = \mathcal{Z}(\|\mathbf{v}_{\text{diff}}(x)\|_2) \times \frac{\|\mathbf{v}_{\text{diff}}(x) - \mathbf{c}\|_1}{\mu_{\text{dist}}}
\end{equation}
where $\mathcal{Z}(\cdot)$ is $Z$-score standardization and $\mu_{\text{dist}}$ is the mean Manhattan distance. As shown in Figure~\ref{fig:geometry}c, PI effectively pushes extremes to the tails, amplifying the mining efficiency for our top/bottom-K sampling.

\paragraph{Rubric Synthesis from Polarized Samples}
We feed the mined samples into a teacher model and distill rubrics via Ensemble Distillation (see Figure~\ref{fig:prompt_stage1} and \ref{fig:prompt_stage2} in Appendix~\ref{app:prompts}). The synthesized rubric serves as a standardized baseline that aligns human experts on a common scoring standard, maximizing Inter-Annotator Agreement; operations teams further inject domain logic, so the final supervision remains human-aligned. This one-off initialization is decoupled from the runtime scoring loop and incurs no marginal cost per session; it is rerun only when the business scenario or backbone changes.

\subsection{Geometry-Aware Regression Probing}
\label{sec:regression}

With the bootstrapped rubric in hand, the remaining task is to map a session's hidden state to a calibrated 1--5 score via four steps.

\paragraph{Independent Human Annotation}
Human experts score each session against the bootstrapped rubric. To measure reliability on the 1--5 ordinal scale, we use \textbf{Krippendorff's alpha (K-$\alpha$)} \citep{krippendorff2018content}, a chance-corrected coefficient that penalizes disagreements by magnitude. The experts reach a strong inter-annotator agreement of K-$\alpha > 0.90$ on the 700-session training set. These labels are the sole supervision signal for the PLS head.

\paragraph{Refined Extraction \& Resampling}
The Blind Probing prompts of Sec.~\ref{sec:bootstrapping} were deliberately rubric-agnostic; with the rubric now finalized, we re-extract $\mathbf{v}_{\text{diff}}$ using detailed rubric prompts (see Figure~\ref{fig:prompt_refined_probe} in Appendix~\ref{app:prompts}) to sharpen the satisfaction direction. A second issue then arises: median scores dominate the training set, threatening regression-to-the-mean collapse. We counter this with Inverse Density Sampling based on Manhattan distance: down-sampling high-density neutral regions and up-sampling low-density tails. This reshapes the training distribution, preventing model collapse (validated in Sec.~\ref{sec:ablation}).

\paragraph{PLS Regression \& Feature Augmentation}
Unsupervised methods such as PCA select directions of maximum \emph{variance}, which on LLM hidden states often track topic rather than satisfaction. We therefore use supervised PLS regression, which selects directions of maximum \emph{covariance} with the target score, filtering topic noise.
To further boost small-sample performance, we concatenate the Norm ($\|\mathbf{v}_{\text{diff}}\|_2$) and the aforementioned Distance ($D_{\text{centroid}}$) with the latent vector: $\tilde{\mathbf{v}} = [\mathbf{v}_{\text{diff}}; \|\mathbf{v}_{\text{diff}}\|_2; D_{\text{centroid}}]$, allowing the linear model to capture non-linear intensity signals (Figure~\ref{fig:geometry}a). With $K{=}5$ components on 700 annotations, the head achieves strong sample efficiency without overfitting.

\paragraph{Uncertainty \& Tiered Evaluation}
We train separate PLS heads on an intermediate layer (Layer 15) and the final layer (Layer 40), defining the Uncertainty Score as $\mathcal{U}(x) = |S_{\text{final}}(x) - S_{\text{mid}}(x)|$. A high $\mathcal{U}$ indicates cognitive conflict and serves as an unsupervised error proxy (validated in Sec.~\ref{sec:uncertainty}). This enables tiered deployment: BoRP scores 100\% of traffic, while only high-$\mathcal{U}$ samples are routed to expensive generative or human review.

\subsection{High-Throughput Inference Engine}
\label{sec:system}

Full-traffic monitoring is only feasible when per-sample probing cost stays within tight budgets. We optimize the dominant easy path (Figure~\ref{fig:teaser} Right) along three axes.

\paragraph{Suffix-Only Probing}

Each session is probed against $N$ rubrics sharing the same dialogue prefix. We leverage SGLang's RadixAttention \citep{zheng2024sglang} to compute the shared prefix once and reuse its KV cache across all $N$ suffixes, reducing complexity from $\mathcal{O}(L_{\text{prefix}} \cdot L_{\text{total}})$ to $\mathcal{O}(L_{\text{prefix}} + N \cdot L_{\text{suffix}})$.

\paragraph{Compression \& Hybrid Deployment}
Long or topic-shifting sessions are first sliced via \emph{intent-shift segmentation}; within each sub-session, ``middle-out'' compression (i.e., retaining the head and tail of the dialogue history while discarding the middle) keeps the Target Response intact. This cuts prefill cost by $\sim$64\%; furthermore, we verified that it strictly preserves the model-human K-$\alpha$ agreement and occasionally yields a slight lift, as truncating the noisy middle turns acts as an effective attention filter. We deploy $>$10B models in AWQ-Int4 and $<$10B models in native BF16 to balance memory and throughput.
\section{Experiments}
\label{sec:experiments}

We validate BoRP across four dimensions: Accuracy (RQ1), Generalization (RQ2), Efficiency (RQ3), and Robustness Analysis (RQ4).

\subsection{Experimental Setup}

\paragraph{Datasets} (i)~\textbf{Industrial}: 700 production sessions with expert 1--5 labels (K-$\alpha > 0.90$). (ii)~\textbf{HelpSteer2} \citep{wang2024helpsteer2}: Multi-turn subset aggregated to session-level scores. To prevent K-$\alpha$ bias from label skew, we evaluate each dimension using balanced subsets (varying counts per dimension, see Appendix~\ref{app:dataset_details}) sampled from official splits to ensure a uniform score distribution. For Industrial, we use 5-fold CV; for HelpSteer2, we evaluate on the official test split. 

\paragraph{Baselines}
(i)~Generative LLM-as-a-Judge: Qwen3-14B and the proprietary Qwen3-Max,\footnote{To comply with data residency, we exclude models requiring cross-border transmission (e.g., GPT-5).} with prompt strategies in Appendix~\ref{app:gen_baselines};
(ii)~M-Prometheus (14B) \citep{pombal2025mprometheus}, a Qwen2.5-14B-Instruct specialized judge with Chinese support (Appendix~\ref{app:prometheus_setup}). We also report BoRP on the same Qwen2.5-14B backbone to isolate paradigm contribution from backbone capacity.

\paragraph{Implementation}
BoRP uses Qwen3-14B/8B with $n{=}5$ PLS components ($\sim$140 samples each). All experiments run on a single A100 (80GB); details in Appendix~\ref{app:implementation}.

\paragraph{Metrics} We evaluate 1--5 scale alignment using \textbf{K-$\alpha$} (defined in Sec.~\ref{sec:regression}), which is more rigorous than exact-match accuracy for ordinal data. \textbf{Pearson} ($r$) is also reported to measure linear association.

\subsection{Main Results (RQ1)}

\begin{table}[h]
\centering
\small
\begin{tabular}{l c c}
\toprule
\textbf{Method} & \textbf{K-$\alpha$} & \textbf{Pearson} \\
\midrule
\multicolumn{3}{l}{\textit{Specialized Judge}} \\
M-Prometheus (Qwen2.5-14B) & 0.49{\scriptsize$\pm$.06} & 0.53{\scriptsize$\pm$.05} \\
\midrule
\multicolumn{3}{l}{\textit{Generative Baselines}} \\
Gen (Qwen3-14B) & 0.62{\scriptsize$\pm$.03} & 0.62{\scriptsize$\pm$.04} \\
Gen (Qwen3-Max) & 0.73{\scriptsize$\pm$.02} & 0.74{\scriptsize$\pm$.02} \\
\midrule
\multicolumn{3}{l}{\textit{Ours (BoRP)}} \\
BoRP (8B) & 0.73{\scriptsize$\pm$.03} & 0.73{\scriptsize$\pm$.02} \\
BoRP (Qwen2.5-14B) & 0.69{\scriptsize$\pm$.02} & 0.70{\scriptsize$\pm$.02} \\
\textbf{BoRP (Qwen3-14B)} & \textbf{0.80}{\scriptsize$\pm$.03} & \textbf{0.81}{\scriptsize$\pm$.03} \\
\bottomrule
\end{tabular}
\caption{\textbf{Main Results on Industrial Dataset (5-fold CV).} Under identical Qwen2.5-14B backbone, BoRP raises K-$\alpha$ from 0.49 to 0.69, isolating the paradigm contribution.}
\label{tab:main_results}
\end{table}

Table~\ref{tab:main_results} highlights three findings.
(i)~\textit{Paradigm-isolated comparison}: on the same Qwen2.5-14B backbone, BoRP raises K-$\alpha$ from 0.49 to 0.69 ($+$0.20), isolating the gain to the judging paradigm (Appendix~\ref{app:gen_baselines}).
(ii)~\textit{OOD fragility of specialized SFT}: M-Prometheus (0.49) trails even untrained Gen-14B (0.62), reflecting domain drift from its English-centric training data and prompt-protocol sensitivity (detailed in Appendix~\ref{app:gen_baselines}).
(iii)~\textit{Evolvability}: a backbone swap (Qwen2.5$\rightarrow$Qwen3) raises BoRP from 0.69 to 0.80 with the same 700 labels; even at 8B scale BoRP matches Gen-Max (0.73 vs.\ 0.73) while outscoring the 14B specialized judge (0.73 vs.\ 0.49).
All pairwise gaps exceed one standard deviation across folds.

\subsection{Generalization Analysis (RQ2)}
We test generalization across four HelpSteer2 dimensions (Table~\ref{tab:helpsteer}), retraining only the PLS head. BoRP defaults to Qwen3; under identical Qwen2.5-14B backbones, BoRP$_{Q2.5}$ exceeds M-Prom on 3/4 dimensions in K-$\alpha$ (Pearson in Appendix~\ref{app:helpsteer_pearson}).

\begin{table}[h]
\centering
\small
\setlength{\tabcolsep}{4pt}
\begin{tabular}{l c c c c}
\toprule
\textbf{Method} & \textbf{Verbosity} & \textbf{Correct.} & \textbf{Coherence} & \textbf{Complex.} \\
\midrule
Gen-14B            & 0.33 & 0.13 & 0.22 & 0.29 \\
Gen-Max            & 0.15 & 0.40 & 0.39 & 0.76 \\
M-Prom             & 0.24 & 0.28 & \underline{0.52} & 0.61 \\
BoRP$_{Q2.5}$      & 0.65 & 0.36 & 0.47 & \textbf{0.78} \\
\textbf{BoRP}      & \textbf{0.70} & \textbf{0.50} & \textbf{0.54} & 0.75 \\
\bottomrule
\end{tabular}
\caption{\textbf{HelpSteer2 K-$\alpha$.} BoRP$_{Q2.5}$ acts as a same-backbone control against M-Prom. \textbf{Bold}: best. \underline{Underlined}: only same-backbone cell where M-Prom leads.}
\label{tab:helpsteer}
\end{table}

\paragraph{Verbosity} BoRP$_{Q2.5}$ scores 0.65 vs.\ M-Prom's 0.24. This 0.41 gap strongly suggests that verbosity bias is paradigm-intrinsic and robust to specialized SFT. Generative baselines suffer from score inflation (Gen-Max) or central tendency (Gen-14B); direct regression avoids both. 

\paragraph{Coherence} Our default BoRP hits the highest K-$\alpha$ (0.54), but M-Prom slightly edges out BoRP$_{Q2.5}$ (0.52 vs.\ 0.47). We attribute this to Coherence requiring explicit step-by-step logical deduction (which Prometheus generates via critiques), whereas BoRP relies on holistic representations. Thus, regression probing excels at neutralizing stylistic biases, but generative evaluators remain complementary for explicit reasoning (Sec.~\ref{sec:limitations}).

\subsection{Efficiency Analysis (RQ3)}
\label{sec:efficiency}

Table~\ref{tab:efficiency} benchmarks inference on a single A100 (80GB). Since BoRP (8B) already achieves near-SOTA accuracy (K-$\alpha$ $\approx$ 0.73, Table~\ref{tab:main_results}), we use it to highlight the model-downsizing advantage against the standard 14B generative configurations.
BoRP (8B) achieves 57,800 sessions/hour---a 9.3$\times$ speedup over M-Prometheus and 7.9$\times$ over Gen (14B)---at merely \$4 per 100K sessions ($>$30$\times$ cheaper than the Qwen3-Max API; Appendix~\ref{app:cost_estimation}).

\begin{table}[h]
\centering
\small
\resizebox{\columnwidth}{!}{%
\begin{tabular}{l c c c}
\toprule
\textbf{Method} & \textbf{Throughput} & \textbf{Task Cost (\$)} & \textbf{Relative Cost} \\
\midrule
Qwen3-Max (API) & N/A & 124 & 100\% (Baseline) \\
Gen (14B) & 7,300 & 28 & 22.6\% \\
M-Prometheus (14B) & 6,200 & 34 & 27.4\% \\
\textbf{BoRP (8B)} & \textbf{57,800} & \textbf{4} & \textbf{3.2\%} \\
\bottomrule
\end{tabular}%
}
\caption{\textbf{Efficiency Comparison.} Throughput in sessions/hour on one A100; Task Cost for 100k sessions (Appendix~\ref{app:cost_estimation}). BoRP (8B) is $\sim$9$\times$ faster and more accurate (K-$\alpha$ 0.73 vs.\ 0.49) than the 14B specialized judge.}
\label{tab:efficiency}
\end{table}

\subsection{Robustness and Ablation Analysis (RQ4)}
\label{sec:analysis}

\paragraph{Ablation Studies}
\label{sec:ablation}

\begin{table}[h]
\centering
\small
\resizebox{\columnwidth}{!}{%
\begin{tabular}{l c c}
\toprule
\textbf{Setting} & \textbf{K-$\alpha$} & \textbf{Pearson} \\
\midrule
\multicolumn{3}{l}{\textit{(a) Training Strategy (on Industrial Dataset)}} \\
\textbf{BoRP (Full)} & \textbf{0.80} & \textbf{0.81} \\
\quad w/o Geometric Resampling & 0.62 & 0.69 \\
\quad w/o PLS (use PCA) & 0.55 & 0.58 \\
\midrule
\multicolumn{3}{l}{\textit{(b) Rubric Source (on HelpSteer2)}} \\
Gen (Max) + Original Rubric & 0.15 & 0.62 \\
Gen (Max) + Bootstrapped Rubric & \textbf{0.22} & \textbf{0.66} \\
\textbf{BoRP (14B) + Original} & \textbf{0.70} & \textbf{0.70} \\
\textbf{BoRP (14B) + Bootstrapped} & \textbf{0.67} & \textbf{0.67} \\
\bottomrule
\end{tabular}%
}
\caption{Comprehensive Ablation Study. (a) Geometric resampling and PLS are essential. (b) Bootstrapped rubrics achieve near-expert performance.}
\label{tab:ablation}
\end{table}

We dissect BoRP's components on the Industrial Dataset (training strategy) and HelpSteer2 (rubrics).
As shown in Table~\ref{tab:ablation}(a), Geometric Resampling is critical: training on the natural long-tail distribution causes K-$\alpha$ to collapse from 0.80 to 0.62, as PLS otherwise regresses to the dominant neutral mean. Replacing PLS with unsupervised PCA further degrades it to 0.55.
Table~\ref{tab:ablation}(b) validates the bootstrapped rubric: it matches expert-curated performance (0.67 $\approx$ 0.70), proving its cold-start utility. Interestingly, for Gen (Qwen3-Max), AI-generated rubrics even outperform human ones (0.22 $>$ 0.15; Appendix~\ref{app:helpsteer_rubric}). Bootstrapping samples were strictly excluded from the test set.

\begin{figure}[t] 
    \centering
    \includegraphics[width=0.95\columnwidth]{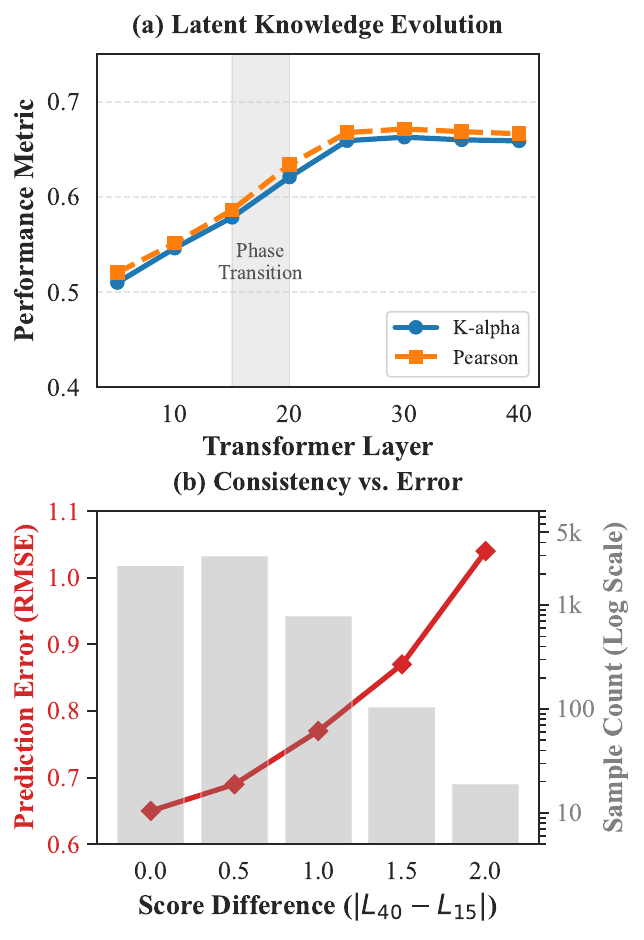}
    \caption{\textbf{Uncertainty Analysis.} (a) Performance stabilizes after Layer 25. (b) Inconsistency correlates with RMSE.}
    \label{fig:uncertainty}
\end{figure}

\paragraph{Uncertainty as an Error Proxy}
\label{sec:uncertainty}
As Figure~\ref{fig:uncertainty}(a) shows that alignment performance stabilizes after Layer 25, we train an intermediate head at Layer 15 and test $\Delta S = |S_{40} - S_{15}|$ as an error proxy on HelpSteer2. Figure~\ref{fig:uncertainty}(b) confirms a strong positive correlation with RMSE: high-consistency samples ($\Delta S \approx 0$) achieve RMSE $\approx$ 0.65, whereas high-conflict samples ($\Delta S > 1.5$) exhibit significantly higher error, validating $\Delta S$ as a zero-cost filter for unreliable predictions.

\paragraph{Boosting A/B Testing Efficiency}
BoRP's calibrated continuous scores compose naturally with standard variance-reduction techniques. Substituting them for the prior binary satisfaction heuristic inside CUPED \citep{deng2013improving} raised the pre-experiment covariate correlation $\rho$, yielding a further $4.5\%$ variance reduction in our online A/B pipeline. The combination reliably detects micro-lifts that binary metrics miss, accelerating iteration cycles.
\section{Conclusion}
We introduced \textbf{BoRP}, an end-to-end ``read, don't write'' framework for industrial LLM evaluation. By projecting a session's hidden state through a lightweight PLS head, BoRP achieves stronger human alignment than generative judges (K-$\alpha$ 0.80 vs.\ 0.73), $>$30$\times$ lower inference cost, and backbone evolvability---all from $\sim$700 expert annotations.
\section{Limitations}
\label{sec:limitations}

We outline five limitations of BoRP.

\paragraph{Modality scope.} The current implementation targets text-only chat; voice scenarios involving acoustic features are not yet covered.

\paragraph{Reasoning depth.} BoRP probes the final hidden states of the base model. For tasks requiring complex logical deduction (e.g., code debugging), if the base model fails to encode these nuances into its final hidden states, a linear probe may struggle to recover them, and explicit Chain-of-Thought judges can be more appropriate.

\paragraph{Structure sensitivity.} The discriminative power of latent extraction depends on feedback signals carried by the dialogue itself. Effectiveness may diminish in single-turn interactions lacking explicit feedback, or in extreme long-tail sessions with frequent topic shifts.

\paragraph{Deployment dependency.} BoRP's advantages in data efficiency, throughput, and backbone evolvability hinge on having access to the base model's hidden states. This is well aligned with on-premise industrial deployment, but precludes pure black-box API settings, where specialized SFT judges may be preferable.

\paragraph{Evaluation vs. Training.} While BoRP provides high-fidelity satisfaction scores for monitoring and A/B testing, it is not a direct replacement for Reward Models (RMs) used in RLHF. Our framework is optimized for session-level absolute scoring rather than providing the dense, step-wise relative preference signals required for policy optimization.

\section*{Ethical Considerations}

\paragraph{Data Privacy \& Anonymization}
The industrial datasets utilized in this study were derived from production logs. To strictly uphold user privacy and comply with data protection regulations (e.g., GDPR), all Personally Identifiable Information (PII) was rigorously anonymized via automated scrubbing pipelines prior to any model exposure. Furthermore, our experimental design strictly adheres to regional data sovereignty laws, ensuring that no internal data was transmitted across borders to third-party API services.

\paragraph{Bias Mitigation}
We acknowledge that the bootstrapped rubrics, being distilled from a Teacher LLM (Qwen3-Max), may inherit intrinsic biases. To mitigate this, our pipeline enforces a Human-in-the-Loop verification step (Sec.~\ref{sec:bootstrapping}) where domain experts review and refine the generated criteria, ensuring they align with ethical standards and do not propagate harmful stereotypes.

\paragraph{Environmental Impact}
A core motivation of BoRP is to democratize high-fidelity evaluation while minimizing energy consumption. By enabling an 8B model to match the performance of closed-source giants, our approach significantly reduces the carbon footprint associated with large-scale continuous monitoring, advocating for sustainable and Green AI practices in the industry.

\bibliography{custom}
\clearpage
\appendix
\onecolumn 
\section{Prompt Templates}
\label{app:prompts}
We provide the English-translated versions of the prompts.
Note: We include empty <think> tags to align with Qwen3's specific instruction-tuning format, ensuring the hidden states are extracted from a valid distribution.
\begin{tcblisting}{colback=white, colframe=gray, title=Blind Probe Template, listing only, breakable}
<|im_start|>system
You are a professional expert in dialogue quality evaluation.<|im_end|>
<|im_start|>user
### Dialogue for Evaluation
```
{Conversation}
```
### Evaluation Dimension
To what extent does the user's behavior or speech express approval of the model's response?

Please describe the performance of this dialogue on the above dimension using a single word.<|im_end|>
<|im_start|>assistant
<think>
</think>
Excellent/Terrible
\end{tcblisting}
\captionof{figure}{The generic prompt used for unsupervised mining (Phase 1) and blind probing.}
\label{fig:prompt_blind_probe}


\begin{tcblisting}{colback=white, colframe=gray, title=Phase 2: Rubric Generation (Stage 1: Case-to-Draft), listing only, breakable}
# Role
You are an expert in dialogue assessment, specializing in distilling abstract, actionable scoring rubrics from concrete conversation examples.
# Task
I need you to create a detailed 1-5 point scoring rubric for the dimension [{rubric_name}].
I cannot explicitly tell you what the standard is, but I will provide two sets of real conversation cases as references.
IMPORTANT: The core object of assessment is "the model's response in the last turn of the multi-turn conversation." Please evaluate the final response within the context of the full dialogue.
1. [High-Score Group]: Conversations that represent the high end (5 points) of the spectrum for this dimension.
2. [Low-Score Group]: Conversations that represent the low end (1 point) of the spectrum for this dimension.
Please compare these two groups, analyze the key differences, and derive a complete 1-5 scoring rubric.
# Input Data
## [High-Score Group] (Represents 5 points)
---
Case High-1:
{good_case_1}
---
Case High-2:
{good_case_2}
---
Case High-3:
{good_case_3}
---
## [Low-Score Group] (Represents 1 point)
---
Case Low-1:
{bad_case_1}
---
Case Low-2:
{bad_case_2}
---
Case Low-3:
{bad_case_3}
---
# Analysis Steps (Chain of Thought)
Please think deeply first and analyze the following questions (do not output the tedious process, just integrate the insights into the final result):
1. Commonality Extraction: In the high-score cases, what are the defining characteristics of the last turn response? Why does it align with the high end of [{rubric_name}]? In the low-score cases, what distinct traits or behaviors appeared in the last turn response that place it at the low end?
2. Feature Localization: What characteristic patterns (e.g., logical structure, specific wording, information depth, formatting, tone) are distinctly different between the two groups?
3. Intermediate State Deduction: If 5 points represents the "maximum presence or adherence" to the high-score traits, and 1 point represents the "minimum presence or opposing traits," what should 3 points look like? (Interpolate the middle ground between the two extremes).
# Output Format
Please directly output the final scoring rubric in the following format:
## [{rubric_name}] Scoring Rubric
### Core Definition
(Summarize the assessment focus of this dimension regarding the last turn response in one sentence)
### Scoring Criteria
- 5 Points (High): [Description of the high extreme] (Signal features: ...)
- 4 Points: [Description] (Signal features: ...)
- 3 Points (Mid): [Deduced intermediate state] (Signal features: ...)
- 2 Points: [Description] (Signal features: ...)
- 1 Points (Low): [Description of the low extreme] (Signal features: ...)
### Key Signals
- High-Score Indicators: ...
- Low-Score Indicators: ...
\end{tcblisting}
\captionof{figure}{The detailed prompt for distilling rubrics from extreme cases.}
\label{fig:prompt_stage1}

\begin{tcblisting}{colback=white, colframe=gray, title=Phase 2: Rubric Generation (Stage 2: Fusion), listing only, breakable}
# Role
You are a senior expert in developing assessment standards. Your task is to integrate multiple draft standards derived from different case studies into a final, authoritative scoring rubric.
# Task
For the dimension [{rubric_name}], we conducted three independent case analyses and obtained three slightly different scoring standards (Drafts).
These drafts may have different focuses or contain biases due to the specificity of the cases used.
Please synthesize these three drafts, extract their core commonalities, eliminate overly one-sided descriptions, and transform concrete scenario-based descriptions into universal abstract descriptions, to output the most universal and robust final version of the scoring rubric.
# Input Drafts
## [Draft V1]
{rule_v1_content}
## [Draft V2]
{rule_v2_content}
## [Draft V3]
{rule_v3_content}
# Output Requirements
1. Deduplication and Abstraction (Critical):
   - If a feature appears in multiple drafts, it is a core feature and must be retained.
   - Generalization: If a draft contains domain-specific examples, you must convert them into universal logic.
     - Bad Example: "Python code execution error" or "Translation grammar error".
     - Good Generalization: "Output contains factual errors or is non-executable".
   - Prohibit Specific Exceptions: The final rubric should not contain specific scenario words unless the dimension itself is explicitly targeted at a specific scenario.
2. Clear Hierarchy: Ensure a distinct step-like progression from 1 to 5 points. The boundaries between adjacent scores (e.g., 3 points vs. 4 points) must be clear, representing a graduated change in the intensity or quality of the attribute being measured.
3. Output Format: Maintain the standard format provided below.
# Final Output (The Master Rubric)
# Evaluation Dimension
- {rubric_name}
# Core Definition
(The synthesized precise definition, applicable to various conversation scenarios)
# Scoring Rubric
- 5 Points (High): [Description of the high end of the spectrum] (Signal features: ...)
- 4 Points: [Description] (Signal features: ...)
- 3 Points (Mid): [Deduced intermediate state] (Signal features: ...)
- 2 Points: [Description] (Signal features: ...)
- 1 Points (Low): [Description of the low end of the spectrum] (Signal features: ...)
# Key Signals
- High-Score Indicators: ...
- Low-Score Indicators: ...
\end{tcblisting}
\captionof{figure}{The prompt for fusing multiple rubric drafts into a final standard.}
\label{fig:prompt_stage2}

\begin{tcblisting}{colback=white, colframe=gray, title=Phase 3: Refined Probing Prompt (With Rubric), listing only, breakable}
<|im_start|>user
# Role
You are a professional dialogue quality analyst. Your task is to analyze the final agent response within the provided dialogue based on a strict, specified evaluation dimension and its scoring rubric, in an objective, evidence-driven manner, and provide a quantitative score.
# Task
1.Carefully read and understand the `[Dialogue for Evaluation]` below to establish context.
2.Assign a score: Evaluate the final agent response and provide an integer score between 1 and 5 based on the scoring criteria.
3.Generate a single-line text result according to the `[Output Format]` below.
# Output Format
```
Please strictly follow the format below for output. Your entire response must only contain this single line of plain text.
Format: `[Score]`
Instructions:
-`[Score]` is an integer from 1 to 5.
-You only need to output the score; DO NOT output the reasoning.
-ABSOLUTELY DO NOT include JSON, Markdown code blocks, newlines, or any other superfluous text, symbols, or formatting.
Output Examples
*   Example 1:
3
*   Example 2:
5
```
# Content for Evaluation
## Dialogue for Evaluation
-Evaluation Object: The following is the complete dialogue record. You need to score only the final agent response.
-Dialogue Format: The dialogue uses `User:Agent:` as role identifiers, one sentence per line, alternating.
```
{Conversation}
```
# Evaluation Dimension
{rubric_name}
# Scoring Rubric
{rubric_detail}
# Special Note
Please focus on the main thread of the dialogue and the final outcome. Ignore scattered interludes.
If the user's behavior fits the characteristics of multiple score ranges, please follow the "Outcome-Oriented Principle" (a good ending results in a high score, a bad ending results in a low score).<|im_end|>
<|im_start|>assistant
<think>
</think>
5/1
\end{tcblisting}
\captionof{figure}{The refined prompt used for the final PLS training (Phase 3).}
\label{fig:prompt_refined_probe}

\section{Generative Baselines: Prompts and Sensitivity Analysis}
\label{app:gen_baselines}

We present the three prompting strategies used for generative baselines and their corresponding performance impact.

Table \ref{tab:prompt_ablation} details the performance of generative baselines under different prompting strategies.
\begin{itemize}
    \item \textbf{Gen (Qwen3-14B):} Best performance is achieved with the "Score-Only" strategy (K-$\alpha$ = 0.62). Adding reasoning (CoT) degrades performance, likely due to limited reasoning capacity leading to hallucinated justifications.
    \item \textbf{Gen (Qwen3-Max):} Best performance is achieved with "Score-then-Reason" (K-$\alpha$ = 0.73), confirming that stronger models benefit from explicit reasoning steps.
    \item \textbf{Gen (Qwen2.5-14B):} Best performance is again achieved with the "Score-Only" strategy (K-$\alpha$ = 0.565), with "Reason-then-Score" collapsing to K-$\alpha$ = 0.245---a $-0.320$ drop attributable solely to the output protocol on the same backbone. This is also the backbone and the output protocol used by the M-Prometheus baseline (Sec.~\ref{sec:experiments}), which provides the natural interpretation of its score below.
\end{itemize}
We used the best-performing configuration for each model in the main paper (Table~\ref{tab:main_results}).

\begin{table}[h]
\centering
\small
\begin{tabular}{l l c c}
\toprule
\textbf{Model} & \textbf{Strategy} & \textbf{K-$\alpha$} & \textbf{Pearson} \\
\midrule
Gen (Qwen3-14B) & Reason-then-Score & 0.577 & 0.590 \\
Gen (Qwen3-14B) & Score-then-Reason & 0.587 & 0.612 \\
\textbf{Gen (Qwen3-14B)} & \textbf{Score Only} & \textbf{0.620} & \textbf{0.623} \\
\midrule
Gen (Qwen3-Max) & Reason-then-Score & 0.690 & 0.710 \\
\textbf{Gen (Qwen3-Max)} & \textbf{Score-then-Reason} & \textbf{0.733} & \textbf{0.737} \\
Gen (Qwen3-Max) & Score Only & 0.697 & 0.698 \\
\midrule
Gen (Qwen2.5-14B) & Reason-then-Score & 0.245 & 0.476 \\
Gen (Qwen2.5-14B) & Score-then-Reason & 0.564 & 0.577 \\
\textbf{Gen (Qwen2.5-14B)} & \textbf{Score Only} & \textbf{0.565} & \textbf{0.582} \\
\bottomrule
\end{tabular}
\caption{Performance of generative baselines across different prompting strategies. We selected the bolded configurations for the main comparison. The Qwen2.5-14B rows extend the analysis to the backbone shared with M-Prometheus.}
\label{tab:prompt_ablation}
\end{table}

\paragraph{Connection to the M-Prometheus Baseline.}
M-Prometheus is built on the Qwen2.5-14B-Instruct backbone with the \emph{Reason-then-Score} output protocol (\texttt{Feedback: ... [RESULT] N}). Our sensitivity analysis (Table \ref{tab:prompt_ablation}) shows that on this backbone, the Reason-then-Score protocol is the most fragile, yielding the lowest performance (K-$\alpha$ = 0.245). While M-Prometheus's specialized SFT partially compensates for this protocol-induced loss (raising it to 0.493), it still falls below the untrained \emph{Score-Only} baseline at 0.565. 

Crucially, specialized generative judges like Prometheus are structurally and functionally \textbf{locked} into this specific protocol by their instruction-tuning distribution. Unlike general-purpose models, they cannot be easily evaluated under a \emph{Score-Only} paradigm without significant performance collapse. In contrast, BoRP reaches 0.690 on the same backbone (Table \ref{tab:main_results}) by bypassing verbal protocols entirely. This confirms that the "alignment tax" of generative judges is a structural limitation that BoRP fundamentally sidesteps.

\begin{tcblisting}{colback=white, colframe=gray, title=Strategy A: Score-Only Prompt, listing only, breakable}
# Role
You are a professional dialogue quality analyst. Your task is to analyze the final agent response within the provided dialogue based on a strict, specified evaluation dimension and its scoring rubric, in an objective, evidence-driven manner, and provide a quantitative score.

# Task
1.Carefully read and understand the `[Dialogue for Evaluation]` below to establish context.
2.Assign a score: Evaluate the final agent response and provide an integer score between 1 and 5 based on the scoring criteria.
3.Generate a single-line text result according to the `[Output Format]` below.

# Output Format
```
Please strictly follow the format below for output. Your entire response must only contain this single line of plain text.
Format: `[Score]`
Instructions:
-`[Score]` is an integer from 1 to 5.
-You only need to output the score; DO NOT output the reasoning.
-ABSOLUTELY DO NOT include JSON, Markdown code blocks, newlines, or any other superfluous text, symbols, or formatting.

Output Examples
*   Example 1:
3
*   Example 2:
5
```

# Content for Evaluation
## Dialogue for Evaluation
-Evaluation Object: The following is the complete dialogue record. You need to score only the final agent response.
-Dialogue Format: The dialogue uses `User:Agent:` as role identifiers, one sentence per line, alternating.
```
{Conversation}
```

# Evaluation Dimension
{rubric_name}

# Scoring Rubric
{rubric_detail}

# Special Note
Please focus on the main thread of the dialogue and the final outcome. Ignore scattered interludes.
If the user's behavior fits the characteristics of multiple score ranges, please follow the "Outcome-Oriented Principle" (a good ending results in a high score, a bad ending results in a low score).
\end{tcblisting}
\captionof{figure}{The "Score-Only" prompt. This strategy forces the model to output a single token, minimizing decoding cost but relying heavily on intuition. It proved optimal for Qwen3-14B.}
\label{fig:prompt_score_only}

\begin{tcblisting}{colback=white, colframe=gray, title=Strategy B: Score-then-Reason Prompt, listing only, breakable}
# Role
You are a professional dialogue quality analyst. Your task is to analyze the final agent response within the provided dialogue based on a strict, specified evaluation dimension and its scoring rubric, in an objective, evidence-driven manner, and provide a quantitative score.

# Task
1. Carefully read and understand the `[Dialogue for Evaluation]` below to establish context.
2. Formulate a core rationale: Summarize the direct evidence for your score in one concise sentence, strictly referencing the specific scoring criteria.
3. Assign a score: Provide an integer score between 1 and 5 based on that rationale.
4. Generate a single-line text result according to the `[Output Format]` below.

# Output Format
```
Please strictly follow the format below for output. Your entire response must only contain this single line of plain text.

Format: `[Score] [Rationale]`

Instructions:
- `[Score]` is an integer from 1 to 5.
- Separate the `[Rationale]` and `[Score]` with exactly one space.
- ABSOLUTELY DO NOT include JSON, Markdown code blocks, newlines, or any other superfluous text, symbols, or formatting.

Output Examples
*   Example 1:
5 The response contains multiple paragraphs of unrequested background information and filler words.
*   Example 2:
1 The response is extremely concise and answers the prompt directly without any extra wording.
```

# Content for Evaluation
## Dialogue for Evaluation
-Evaluation Object: The following is the complete dialogue record. You need to score only the final agent response.
-Dialogue Format: The dialogue uses `User:Agent:` as role identifiers, one sentence per line, alternating.
```
{Conversation}
```

# Evaluation Dimension
{rubric_name}

# Scoring Rubric
{rubric_detail}

# Special Note
Please focus on the main thread of the dialogue and the final outcome. Ignore scattered interludes.
If the user's behavior fits the characteristics of multiple score ranges, please follow the "Outcome-Oriented Principle" (a good ending results in a high score, a bad ending results in a low score).
\end{tcblisting}
\captionof{figure}{The "Score-then-Reason" prompt. The model commits to a score before generating an explanation. It proved optimal for Qwen3-Max.}
\label{fig:prompt_score_reason}

\begin{tcblisting}{colback=white, colframe=gray, title=Strategy C: Reason-then-Score (CoT) Prompt, listing only, breakable}
# Role
You are a professional dialogue quality analyst. Your task is to analyze the final agent response within the provided dialogue based on a strict, specified evaluation dimension and its scoring rubric, in an objective, evidence-driven manner, and provide a quantitative score.

# Task
1. Carefully read and understand the `[Dialogue for Evaluation]` below to establish context.
2. Formulate a core rationale: Summarize the direct evidence for your score in one concise sentence, strictly referencing the specific scoring criteria.
3. Assign a score: Provide an integer score between 1 and 5 based on that rationale.
4. Generate a single-line text result according to the `[Output Format]` below.

# Output Format
```
Please strictly follow the format below for output. Your entire response must only contain this single line of plain text.

Format: `[Rationale] [Score]`

Instructions:
- `[Score]` is an integer from 1 to 5.
- Separate the `[Rationale]` and `[Score]` with exactly one space.
- ABSOLUTELY DO NOT include JSON, Markdown code blocks, newlines, or any other superfluous text, symbols, or formatting.

Output Examples
*   Example 1:
The response contains multiple paragraphs of unrequested background information and filler words. 5
*   Example 2:
The response is extremely concise and answers the prompt directly without any extra wording. 1
```

# Content for Evaluation
## Dialogue for Evaluation
-Evaluation Object: The following is the complete dialogue record. You need to score only the final agent response.
-Dialogue Format: The dialogue uses `User:Agent:` as role identifiers, one sentence per line, alternating.
```
{Conversation}
```

# Evaluation Dimension
{rubric_name}

# Scoring Rubric
{rubric_detail}

# Special Note
Please focus on the main thread of the dialogue and the final outcome. Ignore scattered interludes.
If the user's behavior fits the characteristics of multiple score ranges, please follow the "Outcome-Oriented Principle" (a good ending results in a high score, a bad ending results in a low score).
\end{tcblisting}
\captionof{figure}{The standard Chain-of-Thought (CoT) prompt. While theoretically stronger, it degraded performance on smaller models due to reasoning errors.}
\label{fig:prompt_cot}

\clearpage
\section{Prometheus Baseline Configuration}
\label{app:prometheus_setup}

\paragraph{Models.} Our specialized-judge baseline is M-Prometheus 14B (\texttt{Unbabel/M-Prometheus-14B}), built on Qwen2.5-14B-Instruct \citep{pombal2025mprometheus}. We choose M-Prometheus over Prometheus 1 / Prometheus 2 \citep{kim2024prometheus, kim2024prometheus2} for two reasons: (i) it shares the exact Qwen2.5-14B-Instruct backbone with one of our BoRP variants, enabling clean same-backbone controlled comparison; and (ii) it explicitly supports Chinese, avoiding an unfair language mismatch on our internal Chinese industrial dataset.

\paragraph{Rubric Conversion.} BoRP's expert rubric for User Acceptance (Appendix~\ref{app:rubric_private}) and the bootstrapped HelpSteer2 rubric (Appendix~\ref{app:helpsteer_rubric}) are restructured into the Prometheus 5-tier rubric template: a one-sentence \texttt{criteria description} followed by one description per integer score $\in \{1, \dots, 5\}$. The reference answer field is omitted on the industrial dataset (no gold response is available); for consistency we likewise omit it on HelpSteer2.

\paragraph{Inference.} We use greedy decoding (temperature = 0) with \texttt{max\_new\_tokens} = 512, leaving sufficient budget for the verbal feedback expected by the Prometheus output schema (\texttt{Feedback: ... [RESULT] N}). Outputs are parsed with the regex \texttt{[RESULT] (\textbackslash d)}; unparseable outputs (\textless 0.5\% of samples) default to score 3, matching BoRP's fallback policy.

\paragraph{Hardware \& Cost.} Inference is run on a single A100 (80GB), achieving a throughput of 6{,}200 sessions/hour for M-Prometheus 14B (vs.\ 7{,}300 for Gen-14B and 57{,}800 for BoRP-8B; see Table \ref{tab:efficiency}). The slowdown relative to Gen-14B is attributable to the additional output tokens consumed by verbal feedback.

\paragraph{Reproducibility.} Model weight SHA, the exact rubric template, the parsing regex, and the inference scripts are released at the anonymous repository linked in the paper.

\section{HelpSteer2 Pearson Correlations}
\label{app:helpsteer_pearson}

Table~\ref{tab:helpsteer_pearson} reports Pearson correlations across the four HelpSteer2 dimensions, complementing the K-$\alpha$ results in Table~\ref{tab:helpsteer}. The same column-wise patterns hold: BoRP leads on 4/4 dimensions, and BoRP$_{Q2.5}$ exceeds M-Prom on 3/4 dimensions under identical Qwen2.5-14B backbone. Notably, on Verbosity the Pearson gap (Gen-Max 0.622 vs.\ K-$\alpha$ 0.153) exposes the Score Inflation pattern: generative judges may rank correctly yet miscalibrate the absolute score scale, a failure mode BoRP avoids by direct regression.

\begin{table}[h]
\centering
\small
\setlength{\tabcolsep}{4pt}
\begin{tabular}{l c c c c}
\toprule
\textbf{Method} & \textbf{Verbosity} & \textbf{Correct.} & \textbf{Coherence} & \textbf{Complex.} \\
\midrule
Gen-14B            & 0.451 & 0.33 & 0.22 & 0.41 \\
Gen-Max            & 0.622 & 0.43 & 0.46 & 0.77 \\
M-Prom             & 0.558 & 0.28 & 0.53 & 0.66 \\
BoRP$_{Q2.5}$      & 0.682 & 0.37 & 0.56 & \textbf{0.79} \\
\textbf{BoRP}      & \textbf{0.696} & \textbf{0.55} & \textbf{0.57} & 0.76 \\
\bottomrule
\end{tabular}
\caption{HelpSteer2 four-dimensional Pearson correlations; complement to K-$\alpha$ in Table~\ref{tab:helpsteer}. \textbf{Bold}: best per column.}
\label{tab:helpsteer_pearson}
\end{table}

\clearpage
\section{Rubric Comparison on HelpSteer2}
\label{app:helpsteer_rubric}

\begin{center}
\small 
\begin{tabular}{p{0.05\textwidth} p{0.38\textwidth} p{0.52\textwidth}}
\toprule
\textbf{Score} & \textbf{Official Rubric (Human)} & \textbf{Bootstrapped Rubric (AI)} \\
\midrule
\textbf{1} & 
The response is short, to the point, and the most concise it can be. No additional information is provided outside of what is requested by the prompt. & 
The response is maximally concise, delivering only the precise information required to satisfy the user’s request with zero extraneous words, structure, pleasantries, explanations, or formatting unless explicitly mandated. (Signal features: stripped-down phrasing, absence of greetings/closings/acknowledgments, literal compliance with brevity constraints, minimal sentence count, ends immediately after core information is conveyed.) \\
\midrule
\textbf{2} & 
The response is on the shorter side but could still have words, details, and/or text removed before it’s at a bare minimum of what the response is trying to convey. & 
The response is largely direct but contains a small amount of non-functional phrasing—such as a single redundant clause, a polite preamble, or a closing offer of further help—that slightly dilutes maximal efficiency without adding substantive content. (Signal features: core answer preceded or followed by one short filler phrase, very light hedging or acknowledgment, no examples or structural formatting, near-minimal length.) \\
\midrule
\textbf{3} & 
The response isn’t especially long or short given what the prompt is asking of the model. The length is adequate for conveying a full response but isn’t particularly wordy nor particularly concise. & 
The response is moderately concise—fully addressing the request while including minor non-essential elements such as brief contextual reminders, light politeness markers, or one layer of explanatory padding that does not constitute significant redundancy. (Signal features: single paragraph with slight elaboration, minimal transitional phrasing, brief acknowledgments or softening language, no major repetition or structural excess.) \\
\midrule
\textbf{4} & 
The response is on the longer side but could still have more added to it before it is considered fully detailed or rambling. & 
The response is noticeably wordy, containing clearly superfluous content such as minor digressions, unrequested examples, mild repetition, or over-structured formatting that could be removed without impairing the answer’s completeness. (Signal features: multiple sentences or paragraphs with only partial relevance, inclusion of marginally related context, optional guidance or caveats not solicited, slight inflation of scope through elaboration.) \\
\midrule
\textbf{5} & 
The response is particularly lengthy, wordy, and/or extensive with extra details given what the prompt requested from the assistant model. The response can be verbose regardless of if the length is due to repetition and incoherency or if it is due to rich and insightful detail. & 
The response is excessively verbose, significantly exceeding the informational needs of the query through extensive redundancy, tangential explanations, unsolicited detail, or formalized structuring that distracts from core utility. (Signal features: multi-paragraph output for simple requests, repetitive restatements of the same idea, inclusion of general background or pedagogical content not prompted, exhaustive enumeration or listing beyond necessity, use of section headers, bullets, or academic tone where inappropriate.) \\
\bottomrule
\end{tabular}
\label{tab:rubric_comparison_full}
\captionof{table}{\textbf{Rubric Comparison on HelpSteer2 (Verbosity).} We map the original 0-4 scale to our 1-5 training target.}
\end{center}

\clearpage
\section{Rubric for User Acceptance}
\label{app:rubric_private}
The behavior-oriented rubric used for our main experiments (generated via bootstrapping) is detailed in Table \ref{tab:full_rubric}.

\begin{table}[h]
\centering
\small
\begin{tabular}{p{0.1\linewidth} p{0.8\linewidth}}
\toprule
\textbf{Score} & \textbf{Criteria} \\
\midrule

5 & \textbf{Explicit Affirmation:} Clearly gives a positive evaluation or strong praise for the content or quality of the answer. Characteristics: Uses high-energy words like "Awesome," "Thank you very much," "Well summarized," "This information is very useful."\\
4 & \textbf{Logical Continuation/Successful Recovery:} The user does not praise strongly but accepts the answer behaviorally. Includes:
  - In-depth Follow-up: Asking logically related extension questions based on the answer.
  - Task Continuation: Initiating multiple independent but similar tasks in succession.
  - Flaws do not obscure virtues: Despite local corrections or a poor start, the user eventually expresses clear approval.\\
3 & \textbf{Neutral Reception:} No merit or fault, or the main intent is vague. Includes:
  - The user expresses neither positive nor negative views on the answer.
  - Topic Jumping: Random questions with no logical connection.\\
2 & \textbf{Friction/Struggle:} There are explicit obstacles in the main interaction process. Includes:
  - Questioning/Correction: Pointing out errors and requesting rewrites, or proposing specific modifications/additions.
  - Repeating Questions: Repeating the same question because the answer was not obtained.
  - Deterioration: Ending the conversation with negative feedback after a promising start.\\
1 & \textbf{Explicit Negation:} Total denial or emotional confrontation. Characteristics: Insults, sarcasm, declaring it "completely useless."\\
\bottomrule
\end{tabular}
\caption{Bootstrapped rubric for User Acceptance.}
\label{tab:full_rubric}
\end{table}


\section{Dataset Details}
\label{app:dataset_details}

Table~\ref{tab:dataset_counts} provides the sample counts for the balanced HelpSteer2 subsets used for training and evaluating the PLS regression heads across four dimensions. These subsets are sampled from the official multi-turn splits to ensure a uniform score distribution, which is critical for the rigorous calculation of the K-$\alpha$ agreement coefficient.

\begin{table}[h]
\centering
\small
\begin{tabular}{l c c}
\toprule
\textbf{Dimension} & \textbf{Training Samples} & \textbf{Testing Samples} \\
\midrule
Correctness & 1,200 & 150 \\
Coherence   & 500   & 50  \\
Complexity  & 500   & 50  \\
Verbosity   & 1,200 & 346 \\
\bottomrule
\end{tabular}
\caption{Sample counts for balanced HelpSteer2 multi-turn subsets.}
\label{tab:dataset_counts}
\end{table}

\section{Implementation Details}
\label{app:implementation}

\paragraph{Latent Vector Extraction (PyTorch Reference)}
While our production system utilizes SGLang for high-throughput serving, we provide a standalone PyTorch implementation (using HuggingFace Transformers) to facilitate reproducibility. The snippet below demonstrates how to extract the contrastive hidden states.

\begin{tcolorbox}[title=Probe Extraction Snippet, colback=white]
\begin{lstlisting}[language=Python]

import torch
from transformers import AutoTokenizer, AutoModelForCausalLM

model_path = "Qwen/Qwen3-8B"

model = AutoModelForCausalLM.from_pretrained(
    model_path,
    trust_remote_code=True,
    device_map="auto"
)
tokenizer = AutoTokenizer.from_pretrained(model_path, trust_remote_code=True)
model.eval()

prompt_score_1 = "<|im_start|>system\nYou are a professional expert ... <think>\n\n</think>\n\n1"
prompt_score_5 = "<|im_start|>system\nYou are a professional expert ... <think>\n\n</think>\n\n5"

def get_last_token_vector(prompt: str) -> torch.Tensor:
    inputs = tokenizer(prompt, return_tensors="pt").to(model.device)
    with torch.no_grad():
        outputs = model(inputs, output_hidden_states=True)
        last_token_vec = outputs.hidden_states[-1][0, -1, :]
    return last_token_vec

vec_1 = get_last_token_vector(prompt_score_1)
vec_5 = get_last_token_vector(prompt_score_5)
diff_vector = vec_5 - vec_1
\end{lstlisting}
\end{tcolorbox}

\paragraph{Regression Head Configuration}
We construct the input features by concatenating the difference vector (dim=$D_{model}$) with two scalar geometric features (Norm and Distance), resulting in a $(D_{model}+2)$-dimensional vector. 
For Qwen3-14B ($D_{model}=5120$) and Qwen3-8B ($D_{model}=4096$), the inputs are 5122-dim and 4098-dim respectively.
The regressor is implemented using \texttt{sklearn.cross\_decomposition.PLSRegression} with $n\_components=5$ and \texttt{scale=True}.

\paragraph{Quantization Settings}
For production deployment, we adopt a hybrid strategy:
\begin{itemize}
    \item \textbf{Qwen3-14B:} We utilize pre-quantized AWQ-Int4 checkpoints (e.g., \textit{[Anonymous Repository]/Qwen3-14b-int4-awq}). The weights are quantized to 4-bit integers with a group size of 128, loaded directly via the SGLang kernel.
    \item \textbf{Qwen3-8B:} We deploy in native bfloat16 (BF16) precision to avoid dequantization overhead, as it fits within the A100 memory budget without compression.
\end{itemize}

\section{Cost Estimation Details}
\label{app:cost_estimation}

To ensure the reproducibility of the efficiency analysis presented in Table \ref{tab:efficiency}, we provide a detailed breakdown of the calculation methodologies.

\subsection{Evaluation Parameters}
All cost estimations are normalized to a standard workload of $N = 100{,}000$ sessions. The complexity is defined by the following average token statistics from our Industrial Dataset:

\begin{itemize}
    \item \textbf{Average Input Length ($L_{in}$):} 2,700 tokens (including system prompt, history, and query).
    \item \textbf{Average Output Length ($L_{out}$):} 200 tokens for generative models (reasoning + score); effectively $0$ for BoRP.
\end{itemize}

\subsection{Self-Hosted Models (BoRP \& Gen-14B)}
For local models, cost is a function of inference throughput and hardware pricing.

\paragraph{Hardware \& Pricing.} We assume a single NVIDIA A100 (80GB) spot instance:
\begin{equation}
    P_{gpu} = \$2.00 \text{ per hour}
\end{equation}

\paragraph{Calculation Formula.} The total task cost $C_{self}$ is:
\begin{equation}
    C_{self} = \frac{N}{T_{throughput}} \times P_{gpu}
\end{equation}

\paragraph{BoRP Specifics.} For BoRP (8B) with $T = 57{,}800$ sessions/hour:
\begin{equation}
    C_{BoRP} = \frac{100{,}000}{57{,}800} \times 2.00 \approx \$3.46
\end{equation}

\paragraph{Generative Baseline.} For Gen (14B) with $T = 7{,}300$ sessions/hour:
\begin{equation}
    C_{Gen} = \frac{100{,}000}{7{,}300} \times 2.00 \approx \$27.40
\end{equation}

\subsection{Proprietary API Models}
Costs are calculated based on token consumption using public pricing (as of late 2025).

\paragraph{Pricing Standards.}
\begin{itemize}
    \item \textbf{Qwen3-Max:} $P_{in} = \$0.46$, $P_{out} = \$1.80$ (per 1M tokens).
\end{itemize}

\paragraph{Calculation Formula.} The total task cost $C_{api}$ is:
\begin{equation}
    C_{api} = \frac{N}{10^6} \times (L_{in} \cdot P_{in} + L_{out} \cdot P_{out})
\end{equation}

Based on the average lengths $L_{in}$ and $L_{out}$ above, this yields total costs of \$124 (Qwen3-Max).

\end{document}